\documentclass{article}

\usepackage[final]{neurips_2020}

\usepackage[utf8]{inputenc} % allow utf-8 input
\usepackage[T1]{fontenc}    % use 8-bit T1 fonts
\usepackage{hyperref}       % hyperlinks
\usepackage{url}            % simple URL typesetting
\usepackage{booktabs}       % professional-quality tables
\usepackage{amsfonts}       % blackboard math symbols
\usepackage{amsmath}
\usepackage{nicefrac}       % compact symbols for 1/2, etc.
\usepackage{microtype}      % microtypography
\usepackage{graphicx}
\usepackage{bbm}

\title{Visualizing the Loss Landscape of Actor Critic Methods with Applications in Inventory Optimization}

\author{%
  Recep Yusuf Bekci  \\
  Desautels Faculty of Management\\
  McGill University\\
  Montreal, Canada \\
  \texttt{recep.bekci@mail.mcgill.ca} \\
  \And
  Mehmet Gümüş  \\
  Desautels Faculty of Management\\
  McGill University\\
  Montreal, Canada \\
  \texttt{mehmet.gumus@mcgill.ca}
}

\begin{document}

\maketitle

\begin{abstract}
Continuous control is a widely applicable area of reinforcement learning. The main players of this area are actor-critic methods that utilize policy gradients of neural approximators as a common practice. The focus of our study is to show the characteristics of the actor loss function which is the essential part of the optimization. We exploit low dimensional visualizations of the loss function and provide comparisons for loss landscapes of various algorithms. Furthermore, we apply our approach to multi-store dynamic inventory control, a notoriously difficult problem in supply chain operations, and explore the shape of the loss function associated with the optimal policy. We modelled and solved the problem using reinforcement learning while having a loss landscape in favor of optimality.
\end{abstract}

\section{Introduction}

 Reinforcement learning agents explore the state and action space and develop behaviour policies using feedbacks from the environment. The size of the observation and policy space can make the methodology harder, even the most advanced computation systems can suffer from the curse of dimensionality. Approximate solution methods are one of the fundamental building blocks of reinforcement learning. Many reinforcement learning applications in challenging domains are available thanks to the recent developments in function approximation. Utilizing artificial neural networks as non-linear function approximators is a common practice in the literature. Policy gradient methods are examples of the literature that learns parameterized policies in continuous domains.

In this study, we would like to analyze the loss functions of some policy gradient methods. We selected two algorithms on continuous control domain one of which uses deterministic policy and the other uses stochastic policy while having state-of-art performance results on OpenAI tasks. \citep{brockman2016openai} However, they require separate hyperparameter tuning processes for different tasks. \citep{gu2016q} It is also widely known that policy gradient methods are sensitive to hyperparameters and they require a serious hyperparameter tuning process. We modified special parts of the algorithms and examine their effects on loss shape and performance. In this way, we try to explain and visualize the reasons why some model settings are more successful to learn some tasks than others and make use of a visualization methodology in actor-critic method design.

In addition, we modelled a multi-store multi-product inventory control model as a Markov Decision Process. It is an important problem in supply chain operations yet it lacks a holistic approach. We trained a state-of-art agent on the model. Again, we provide loss function plots for this problem. The visualization methodology especially valuable for that kind of problem because convexity is a vital part of characterizing optimal policies in stochastic inventory theory.

\section{Background}

\subsection{Visualizing the Loss Function}
Neural networks are high dimensional models works on empirical risk minimization principle. To make use of neural networks, an appropriate loss function is optimized by adjusting the trainable parameters, also known as weights. The system can be described as a loss function and parameters. The number of parameters makes it impossible to plot the model loss and see curvature with respect to the parameters. We use the method proposed in \cite{goodfellow2014qualitatively} and generalized for three-dimensional plots in \cite{li2018visualizing} as follows
\begin{equation}
    f(\alpha, \beta) = J(\theta^* + \alpha\omega_1 + \beta\omega_2)
\label{method}
\end{equation}
where $\alpha$ and $\beta$ are the reduced dimensions. $\theta^*$ is optimal parameters, and $\omega_1$ and $\omega_2$ are standard Gaussian random vectors of the same size as parameters. Note that, $\omega_1$ and $\omega_2$ are nearly orthogonal with high probability.

Using the representation in Equation \ref{method}, we are applying a huge dimensionality reduction on the loss function. We plot $f(\alpha, \beta)$ versus $[-1,1]$  grids $\alpha$ and $\beta$ to observe neighbourhood of the optimal point. In the resulting 3D graph, the non-convex structure asserts that the full loss function also has non-convexity. On the other hand, a convex 3D graph is only a clue of the low non-convexity of the loss function. The state of the approximator does not necessarily have the optimality property. One can initialize a neural network and visualize the current state of parameters, i.e. random weights. In that case, the visualization still works and shows the neighbourhood of the current part of the loss function. The existence of the cone shape in the plot is a sign of the training phase on the parameters.

In this study, we work on the loss functions of actor approximators. The network structure of our interest is fully connected layers accompanied by ReLU non-linearities. We used layer-wise normalization on $f(\alpha, \beta)$ to neutralize the scale invariance of such architecture. \citep{li2018visualizing}

\subsection{Dynamic Inventory Model for Seasonal Products}
\label{dyninv}
Dynamic inventory management is a well-studied subject in Operations Research. In the general setting, it is modelled as a Markov Decision Process(MDP) for a single product. \cite{karlin1960dynamic} and \cite{scarf1959optimality} showed that there exists an optimal policy for different cost settings. Our focus is on a more specific case where product procurement is done at the beginning of the season and the product is distributed periodically to the stores. This model can be applied to any real-life scenarios that include bulk procurement and periodic distribution. A natural application is in the fashion industry.

Multiple stores and multiple products are the sources of the complexity of the problem. \cite{caro2010inventory} discussed a similar situation where the correlations between stores and products are ignored. They claim of usage of a heuristic algorithm and to best of our knowledge, there is neither any optimality guarantee for the problem nor a holistic modelling approach.

Our model can be defined by tuple $(S,A,P_a,R_a)$ where state space $S$ and action space $A$ are continuous. We model the single store multi product case as follows
\begin{equation}
\label{dyneq}
V_t(q_{it}, I_{it}) = 
\min_{a_{it} \leq q_{it}} \
\mathbb{E}_{\phi} [c_t(a_{it}, I_{it}) + \gamma V_{t+1}(q_{it+1}, I_{it+1}) ]
\end{equation}
The dynamic equation in Equation \ref{dyneq} means that $q_{it}$ and $I_{it}$ are the amount of inventories for depot and store at period $t$ for product $i$. The decision is the flow amount of product $i$ between the depot and the store that is $a_{it}$. Demands are stochastic and determined by probability distribution $\phi(\cdot)$. The equation has two parts, $c_t(a_{it}, I_{it})$ is the immediate cost which occurs in period. We assume that holding inventory at the depot is free of charge. We also have a future cost part accompanied by discounting factor $\gamma$. Inventory level at depot is updated by the action, whereas, demand and action are in charge of the update of the inventory at the store. We assume that backordering is not allowed, namely, if there is no inventory on hand in any period a lost sales cost incurs.

Multi-store multi-depot model can be defined by the following dynamic equation
\begin{equation}
\label{dyneq_multidepot}
V_t(\mathbf{q_t}, \mathbf{I_{t}}) =
\min_{\sum_r a_{irdt} \leq q_{idt} \ \forall d} \
\mathbb{E}_{\mathbf{u} \sim \phi} \Big[\sum_r c_{rt}(a_{irdt}, I_{irt}) + \gamma V_{t+1}(\mathbf{q_{t+1}}, \mathbf{I_{t+1}}) \Big]
\end{equation}
with immediate cost in period $t$ occurs according to
\begin{align}
\label{immcost}
c_{rt}(a_{irdt}, I_{irt}) &= K_{ird}\mathbbm{1}_{a_{irdt}>0} +  W \cdot a_{irdt} \notag \\ &\phantom{{}=1} +  f_{ir}(u_{irt}-I_{irt}-\sum_d a_{irdt})^+  \notag\\ &\phantom{{}=1} +  h_{ir}(I_{irt}+\sum_d a_{irdt}- u_{irt})^+ \
\end{align}

where product $i\in \mathcal{F}$, store $r\in \mathcal{R}$, depot $d\in \mathcal{D}$ and $(x)^+ = max(0, x)$. In each period the elements of immediate cost consist of a fixed cost of ordering $K_{ird}$, a variable cost $W$, a cost of lost sales $f_{ird}$, and a cost of holding inventory at a store $h_{ird}$. The update of inventory of product $i$ at store $r$
\begin{equation}
 I_{irt+1}\leftarrow \Big (I_{irt} + \sum_d a_{irdt} - u_{irt} \Big )^+ \quad\quad  u_{irt}\sim \phi(\cdot)
\end{equation}
and at depot $d$
\begin{equation}
q_{idt+1} \leftarrow q_{idt} - \sum_r a_{irdt}
\end{equation}
After $N$ periods the season ends and a leftover inventory cost occurs according to
\begin{equation}
\label{basecase}
V_N(q_{idN}, I_{irN}) = q_{idN}*s_{ir} + I_{irN}*(h_{ir}+s_{ir})
\end{equation}
where $s_{ir}$ is the salvage cost of products.

The multi-depot model has $|\mathcal{F}|\times|\mathcal{R}|\times|\mathcal{D}|$ dimensions of action space and $|\mathcal{F}|\times|\mathcal{D}| + |\mathcal{F}|\times|\mathcal{R}|$ dimensions of state space. The multi-store case can be modelled and managed with exercising a single depot. This setting performs a high level model while not losing much due to the pooling of possible multiple depots. We prefer to use single depot in this study. The action, $a_{irt}$, specifies the flow of product $i$ to store $r$ in each period. Hence, the action space has $|\mathcal{F}|\times|\mathcal{R}|$ dimensions. Additionally, in each period actions are subject to a constraint defined by
\begin{equation}
\label{multicons}
\sum_r a_{irt} \leq q_{it}
\end{equation}
Immediate cost function and store inventory variables are modified in a similar fashion. Therefore the state dimension of the environment is $|\mathcal{F}| + |\mathcal{F}|\times|\mathcal{R}|$. We can represent the dynamic equation for multi-product multi-store model as follows
\begin{equation}
\label{dyneq_multi}
V_t(\mathbf{q_t}, \mathbf{I_{t}}) =
\min_{\sum_r a_{irt} \leq q_{it}} \
\mathbb{E}_{\phi} \Big[\sum_r c_{rt}(a_{irt}, I_{irt}) + \gamma V_{t+1}(\mathbf{q_{t+1}}, \mathbf{I_{t+1}}) \Big]
\end{equation}
where $\mathbf{q_t}$ and $\mathbf{I_{t}}$ are matrices. Note that the demand distribution requires specific arrangements to build desired correlation between products and stores. We designate demand $\phi(\cdot)$ as Gaussian Process.

Having a large action and observation space is the primary reason for the difficulty of the problem. On top of that, the environment does not necessarily have a stationary distribution of demand, which distinguishes the application of reinforcement learning on inventory management than robotic control.

\section{Experiments}
We conducted two streams of experiments. In the next section, we give loss visualizations of two recently proposed successful off-policy policy gradient algorithms. Soft actor-critic algorithm(SAC) \citep{haarnoja2018soft} utilizes stochastic policy via entropy maximization. The deterministic variant of the algorithm found to be unstable \citep{haarnoja2018soft_2}. We present a comparison of two settings through their loss landscape. Another algorithm in the continuous control domain is Twin Delayed Deep Deterministic Policy(TD3) \citep{fujimoto2018addressing} algorithm which is a Double Q-Learning \citep{van2016deep} variant. TD3 has improvements in value function overestimation through a delay mechanism to reduce the variance of policy updates. In particular, we will focus on the smoothing procedure that TD3 uses during action value calculations.

Moreover, we apply SAC on our multi-store multi-product inventory management environment under various demand scenarios. 

\subsection{MuJoCo Environments}
In this section, we train SAC and TD3 on continuous control tasks of OpenAI Gym \citep{brockman2016openai} as provided in related papers. 

\begin{figure}[h]
\centering
\includegraphics[width=0.9\linewidth]{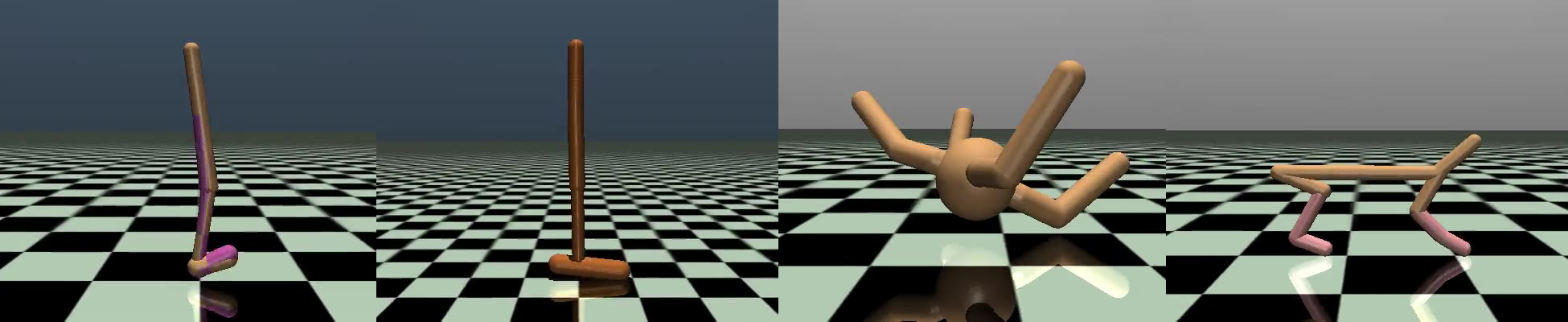}
\caption{Environments. Left to right: Walker2d, Hopper, Ant, HalfCheetah}
\end{figure}

We train models for 1 million episodes in different environments and have similar results claimed in their original papers. Then, we run trained models on related environments to get trajectories. The resulting trajectories have various lengths of states. We apply uniform sampling to get uniform length trajectories and calculate loss values on them.

\begin{figure}[h]
\centering
\includegraphics[width=0.8\linewidth]{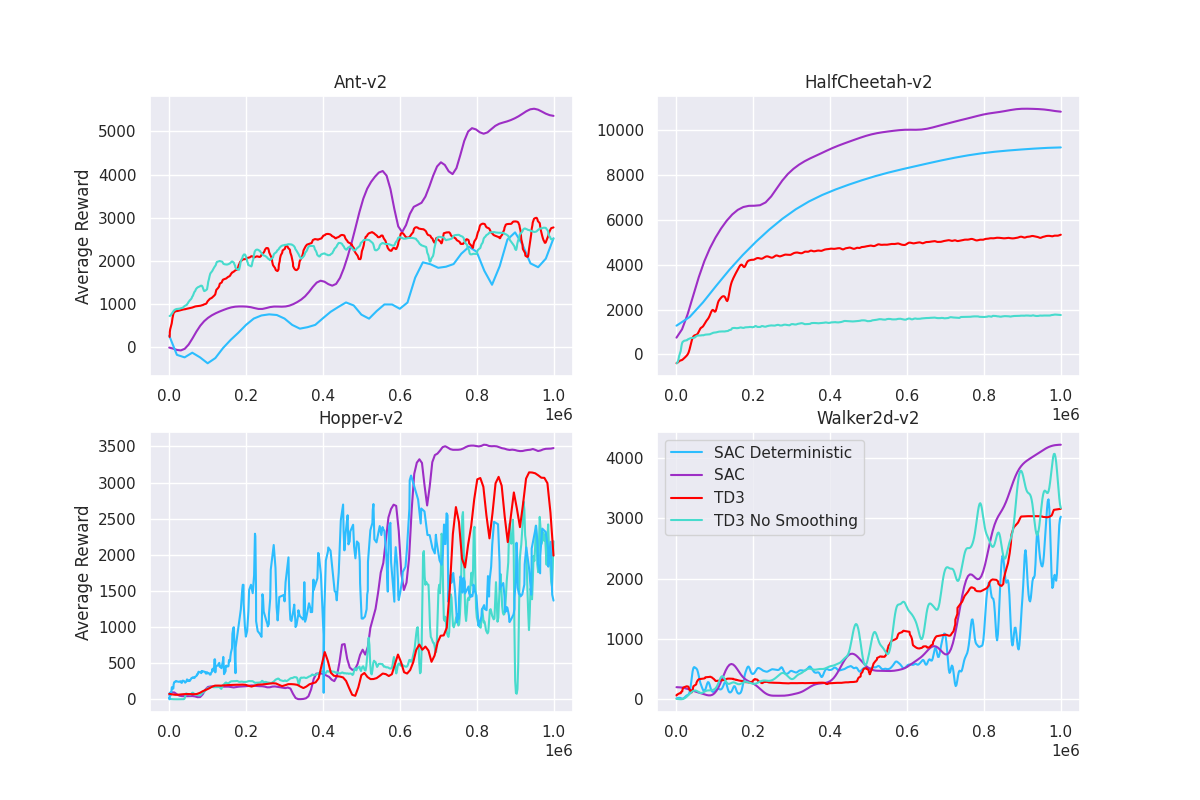}
\caption{Training performances on MuJoCo environments (smoothed)}
\label{mujo}
\end{figure}

We modify the following objective of SAC
\begin{equation}
\label{sacmod}
\pi^* = arg\max_{\pi} \sum_t \mathbb{E}_{s_t,a_t \sim p_{\pi}}  [r(s_t,a_t) + \alpha \mathcal{H}(\pi(\cdot| s_t))]
\end{equation}
and used $\alpha = 0$ to get deterministic version of it while disregarding the entropy part. Note that $\alpha$ is the temperature term that adjusts the importance of $\mathcal{H}(\cdot)$ which is the entropy function. In the original paper SAC is trained for up to 10 million steps however we were able to distinguish performance differences after 1 million. The deterministic version of SAC has slightly worse performance among our four experiments in that category and has stability problems as discussed in the original paper.

\begin{figure}[h]
\centering
\includegraphics[width=0.9\linewidth]{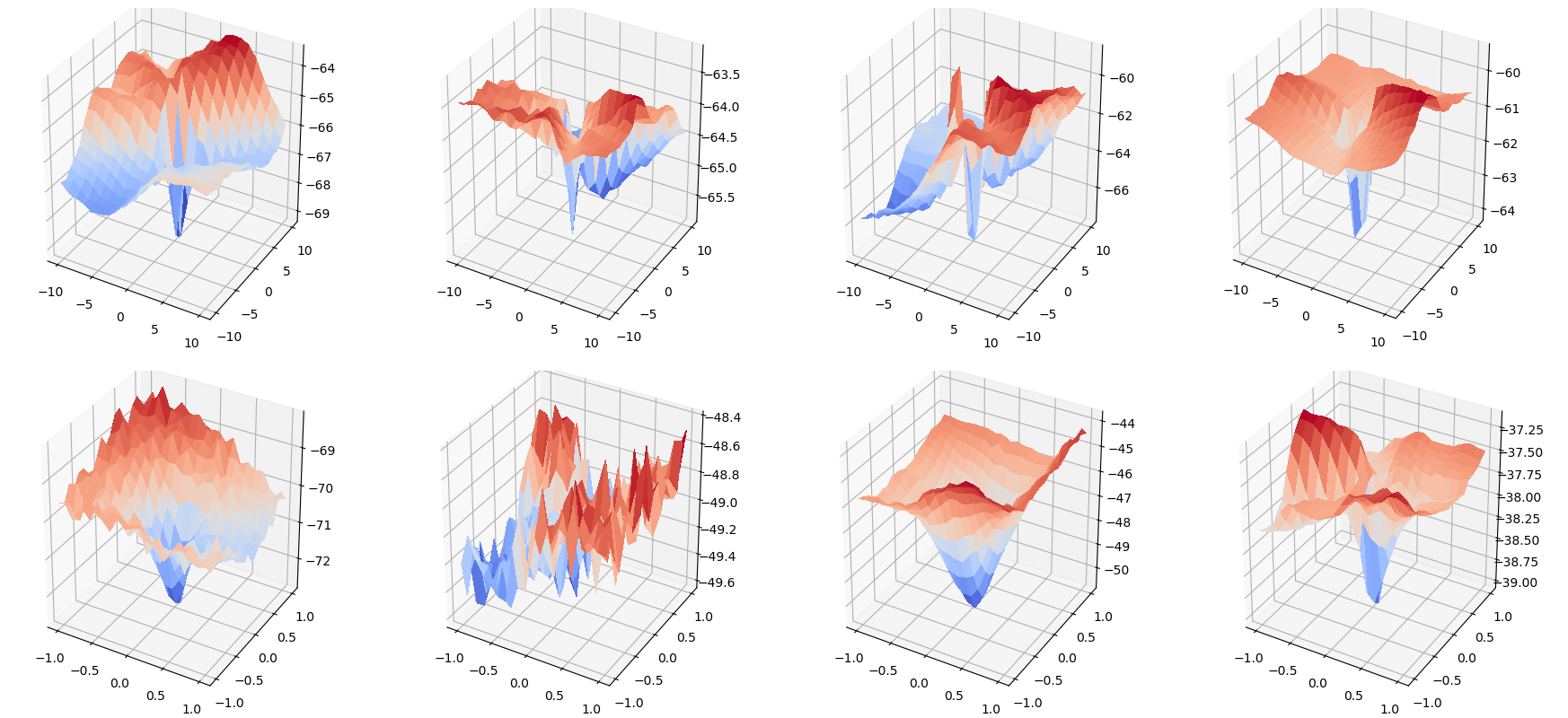}
\caption{Actor loss functions of SAC trained on Walker2d, Hopper, Ant, and HalfCheetah. Stochastic policy on the upper row, deterministic policy on the lower row.}
\label{sac}
\end{figure}

Figure \ref{sac} is quite explanatory for the instability of the deterministic policy compared with stochastic policy. It is very obvious that the loss functions on the four environments are promoting the global minimum while using stochastic policy. Whereas, the deterministic policy resulted in a chaotic loss landscape, especially on $Hopper-v2$. 

We experiment on TD3 with and without action smoothing. In the original paper \citep{fujimoto2018addressing} target value is calculated by
\begin{equation}
\label{actsoom}
y = r + \gamma Q_{\theta'}(s', \pi_{\omega'}(s') + \epsilon)
\end{equation}
where $Q_{\theta'}$ and $\pi_{\omega'}$ are critic and actor networks, $r$ is the reward and $\epsilon$ is clipped Gaussian noise.

\begin{figure}[h]
\centering
\includegraphics[width=0.9\linewidth]{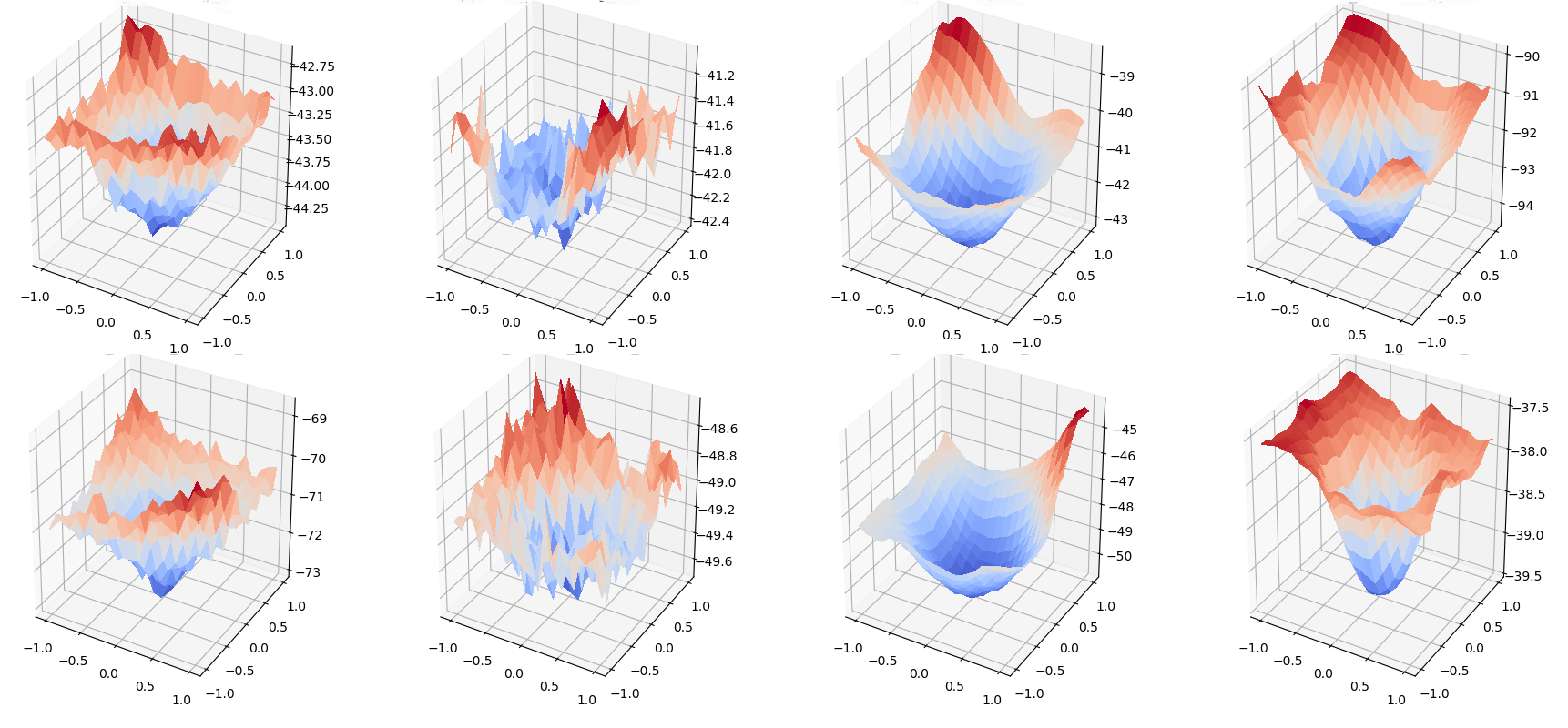}
\caption{Actor loss functions of TD3 trained on Walker2d, Hopper, Ant, and HalfCheetah. Action smoothing on the upper row.}
\label{td3}
\end{figure}

 Without action smoothing the target value is calculated by 
\begin{equation}
\label{actsoom}
y = r + \gamma Q_{\theta'}(s', \pi_{\omega'}(s'))
\end{equation}

Action smoothing in TD3 is not as vital as the stochastic policy for SAC. As a result, we cannot observe a huge performance difference on average rewards. The results in Figure \ref{td3} validate this fact. We can see a little effect of smoothing between the upper and the lower rows of Figure \ref{td3}. In fact, the average rewards of the smoothed algorithm are slightly higher than the one which does not utilize smoothing.

When we compare Figure \ref{sac} and Figure \ref{td3}, we see that the nature of SAC, points the global minimum more harshly even though it is an entropy maximizer algorithm and utilizes soft policy evaluation and soft policy iteration. Note that, SAC over-performs TD3 in every task we analyzed. Hence, we can infer that the comparison of the smoothness of dimensionally reduced loss function in a specific task does not give any idea about the performances of different algorithms. Alternatively, we can think of harsh pointing of global minimum as the effect of a black hole on spacetime which creates strong gravity, therefore a strong force to find the global minimum. We also observe a high variability on $Hopper-v2$ task as a common behaviour. This effect can be inferred from figures \ref{sac} and \ref{td3} where the loss shapes of $Hopper-v2$ are chaotic except the one for SAC.

\subsection{Multi-Product Multi-Store Seasonal Inventory Management}

\begin{figure}[h]
\centering
\includegraphics[width=0.9\linewidth]{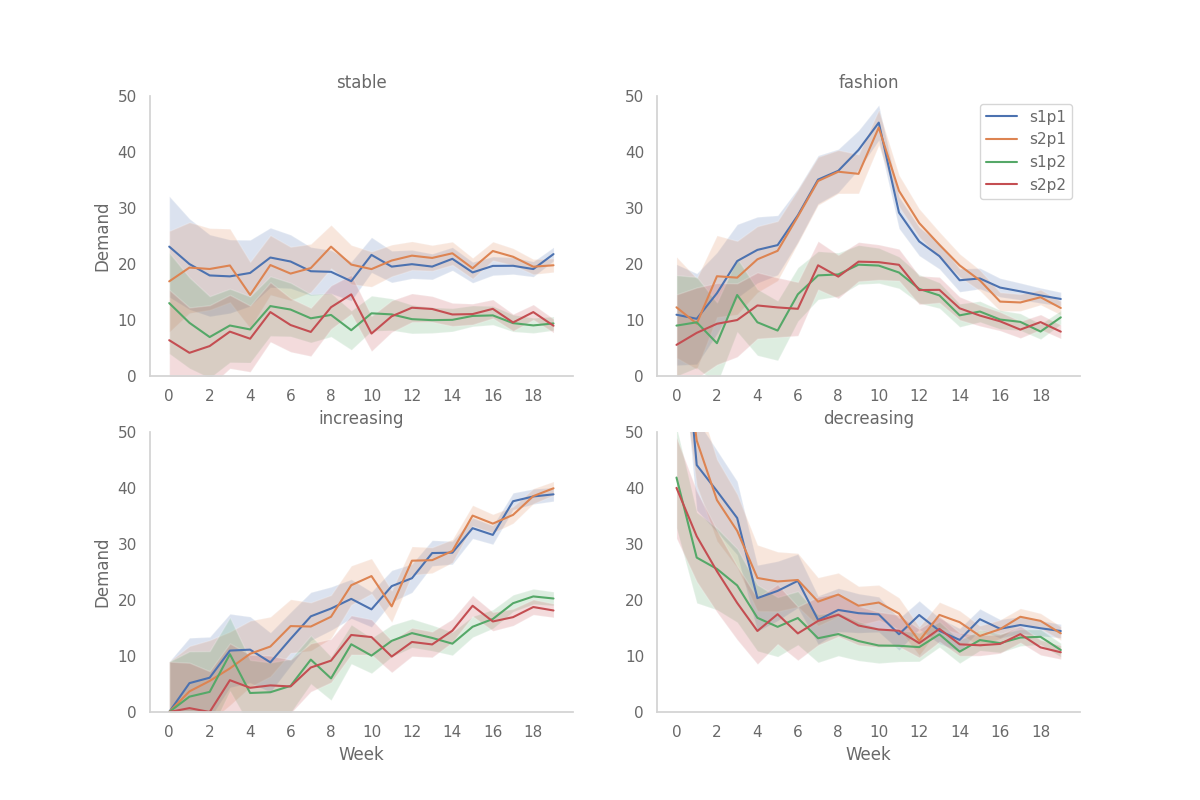}
\caption{Demand distributions of scenarios. $s1p1$ stands for the demand for product 1 in store 1}
\label{demandplot}
\end{figure}

We designed the model in Section \ref{dyninv} as a reinforcement learning environment and use SAC to solve it. We call a full season as one episode. A primary design question is reward design. The problem is cost minimization by its nature. We tested periodic reward versus giving the reward at the end of the episode and decided to use the total cost at the end of the episode.  Intuitively, this corroborates the fact that minimizing the cost of a period is not necessarily the right action to solve the general minimization problem. In addition to that, we scaled the reward and avoid constant negativity of rewards with the help of the following formula
\begin{equation}
\frac{k}{\textrm{Total cost during an episode}}
\end{equation}
where $k$ requires tuning.

We conducted the experiment over 20 periods and 5 different random seeds. We used fixed cost $K=0$, variable cost of ordering $W=0.1$, cost of lost sales $f_{ir}=50$, cost of holding inventory at stores $h_{ir} = 1$, and salvage cost $s_{ir}=10$ for $i\in\{1,2\}$ and $r\in\{1,2\}$.

We modelled demand as Gaussian Process where the products in different stores correlated. We designed four different scenarios to capture different real-world settings. Increasing and decreasing scenarios have monotonic increasing and decreasing demand patterns throughout the episode. Fashion scenario resembles the fashion product life cycle in which demand enjoys growth, maturity, and decline periods respectively. Lastly, a stationary demand scenario does not have any trend as the name suggests. To see the relative performance of our implementation we created two different ordering procedures. The first procedure naively orders the mean demand rate. Moreover, we adapted a base stock policy that keeps the inventory at a designated level of $S$.

% \usepackage{booktabs}
\begin{table}[]
\centering
\caption{Cost improvements of different policies under different scenarios}
\label{tab:my-table}
\begin{tabular}{@{}llllll@{}}
\cmidrule(l){3-6}
                 &                 & \multicolumn{4}{l}{Improvement Over Ordering Mean (\%)} \\ \midrule
Demand Scenarios & Policy          & Mean        & Min         & Max         & Std Dev       \\ \midrule
Decreasing       & Order up to $S$ & 5.68        & -6.42       & 11.44       & 6.61          \\
                 & SAC             & 27.24       & 14.53       & 35.39       & 7.21          \\
Increasing       & Order up to $S$ & 30.67       & 24.26       & 34.05       & 3.64          \\
                 & SAC             & 11.29       & 3.88        & 16.03       & 4.14          \\
Fashion          & Order up to $S$ & 16.19       & 4.82        & 24.89       & 6.84          \\
                 & SAC             & 15.27       & 1.52        & 23.27       & 7.31          \\
Stationary       & Order up to $S$ & 20.28       & 11.26       & 26.67       & 5.42          \\
                 & SAC             & 15.58       & 7.37        & 22.94       & 5.16         
\end{tabular}
\end{table}

We used 2-stores 2-products version of the model in our experiments. However, a real-life application of the problem for operations of a medium-sized global organization can have hundreds of stores and thousands of products which makes the problem interesting and challenging. Also, real-life decisions of promising domains will probably require discrete decisions. We think that modelling and solving the problem in continuous space can be done without diverging from optimality.

Ordering the mean level is the most costly policy as expected. We built our baseline on the mean level ordering policy and presented relative improvements of the other policies. Furthermore, we observe that the policy trained with SAC does not always perform best. Under increasing and stable scenarios, the order-up-to policy yielded a better cost and both algorithms perform similarly under the fashion scenario.

\begin{figure}[h]
\centering
\includegraphics[width=0.9\linewidth]{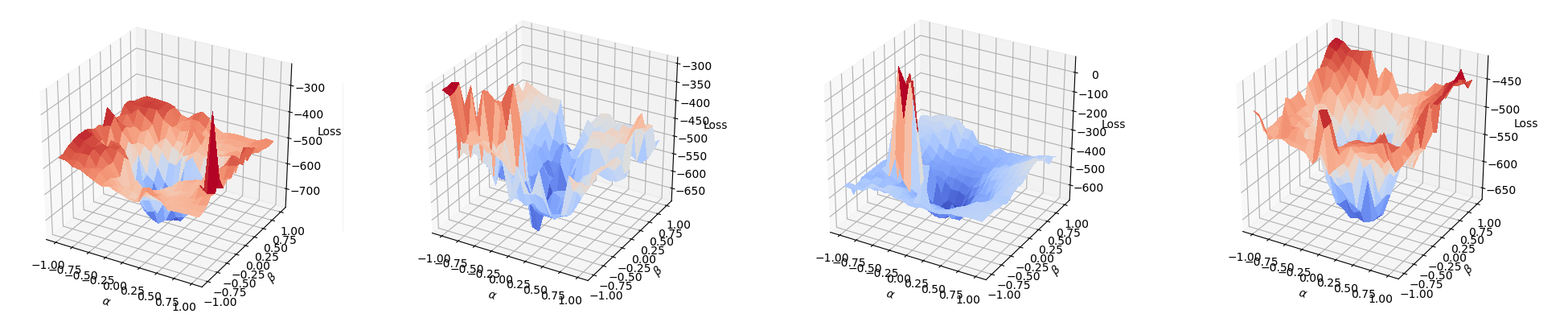}
\caption{Actor loss functions of SAC trained on the dynamic inventory model under decreasing, increasing, fashion, stable demand scenarios(left to right)}
\label{inv_shapes}
\end{figure}

From Figure \ref{inv_shapes}, we can observe that actor losses point out the minimum except for the increasing scenario. Other than increasing scenario, low dimensional representations of loss shapes tend to be convex. Thus, applying SAC on a multi-store multi-product inventory problem has promising results. Since we were able to beat the performance of the trained policy using classical intuitive policies, we can infer that the problem needs more special treatment to solve the problem more efficiently. Our interpretation about the reason behind the inability of SAC to be successful in even smaller 2-stores 2-products model is the nature of the problem which involves a conflict of interests between different products and different stores. The problem cannot be treated independently, besides, it should gain information from correlations of different interests. Therefore, an algorithm which designed for a single goal cannot be successful in that problem area. Possible research directions can include the generalization of continuous control on high dimensions using high-level representations of the space or dedicated algorithm design.

\section{Conclusion}
We applied a visualization technique that can be used to visualize highly parameterized statistical models in reinforcement learning and we presented the action loss landscapes of two actor-critic algorithms and their variants. The results we have, exhibit that the graphs produced using such a technique carries valuable information and can be beneficial for obtaining insights about the algorithms and their applications.  An application of the technique can be tracing the convexity of dynamic programming models. We modelled and solved such a problem in inventory optimization. The results from that specific domain assert that multi-store multi-product dynamic inventory control can be done using reinforcement learning.

\nocite{sutton1998introduction}
\nocite{o2016combining}
\nocite{mnih2015human}
\nocite{barto1983neuronlike}
\nocite{mnih2013playing}
\nocite{porteus2002foundations}
\nocite{liao2017theory}
\nocite{lillicrap2015continuous}

\begin{ack}
We would like to thank Compute Canada for providing computational resources.
\end{ack}

\bibliography{neurips_2020.bbl}

\begin{thebibliography}{}

\bibitem[Barto et~al., 1983]{barto1983neuronlike}
Barto, A.~G., Sutton, R.~S., and Anderson, C.~W. (1983).
\newblock Neuronlike adaptive elements that can solve difficult learning
  control problems.
\newblock {\em IEEE transactions on systems, man, and cybernetics},
  (5):834--846.

\bibitem[Brockman et~al., 2016]{brockman2016openai}
Brockman, G., Cheung, V., Pettersson, L., Schneider, J., Schulman, J., Tang,
  J., and Zaremba, W. (2016).
\newblock Openai gym.
\newblock {\em arXiv preprint arXiv:1606.01540}.

\bibitem[Caro and Gallien, 2010]{caro2010inventory}
Caro, F. and Gallien, J. (2010).
\newblock Inventory management of a fast-fashion retail network.
\newblock {\em Operations Research}, 58(2):257--273.

\bibitem[Fujimoto et~al., 2018]{fujimoto2018addressing}
Fujimoto, S., Van~Hoof, H., and Meger, D. (2018).
\newblock Addressing function approximation error in actor-critic methods.
\newblock {\em arXiv preprint arXiv:1802.09477}.

\bibitem[Goodfellow et~al., 2014]{goodfellow2014qualitatively}
Goodfellow, I.~J., Vinyals, O., and Saxe, A.~M. (2014).
\newblock Qualitatively characterizing neural network optimization problems.
\newblock {\em arXiv preprint arXiv:1412.6544}.

\bibitem[Gu et~al., 2016]{gu2016q}
Gu, S., Lillicrap, T., Ghahramani, Z., Turner, R.~E., and Levine, S. (2016).
\newblock Q-prop: Sample-efficient policy gradient with an off-policy critic.
\newblock {\em arXiv preprint arXiv:1611.02247}.

\bibitem[Haarnoja et~al., 2018a]{haarnoja2018soft_2}
Haarnoja, T., Zhou, A., Abbeel, P., and Levine, S. (2018a).
\newblock Soft actor-critic: Off-policy maximum entropy deep reinforcement
  learning with a stochastic actor.
\newblock {\em arXiv preprint arXiv:1801.01290}.

\bibitem[Haarnoja et~al., 2018b]{haarnoja2018soft}
Haarnoja, T., Zhou, A., Hartikainen, K., Tucker, G., Ha, S., Tan, J., Kumar,
  V., Zhu, H., Gupta, A., Abbeel, P., et~al. (2018b).
\newblock Soft actor-critic algorithms and applications.
\newblock {\em arXiv preprint arXiv:1812.05905}.

\bibitem[Karlin, 1960]{karlin1960dynamic}
Karlin, S. (1960).
\newblock Dynamic inventory policy with varying stochastic demands.
\newblock {\em Management Science}, 6(3):231--258.

\bibitem[Li et~al., 2018]{li2018visualizing}
Li, H., Xu, Z., Taylor, G., Studer, C., and Goldstein, T. (2018).
\newblock Visualizing the loss landscape of neural nets.
\newblock In {\em Advances in Neural Information Processing Systems}, pages
  6389--6399.

\bibitem[Liao and Poggio, 2017]{liao2017theory}
Liao, Q. and Poggio, T. (2017).
\newblock Theory of deep learning ii: Landscape of the empirical risk in deep
  learning.
\newblock {\em arXiv preprint arXiv:1703.09833}.

\bibitem[Lillicrap et~al., 2015]{lillicrap2015continuous}
Lillicrap, T.~P., Hunt, J.~J., Pritzel, A., Heess, N., Erez, T., Tassa, Y.,
  Silver, D., and Wierstra, D. (2015).
\newblock Continuous control with deep reinforcement learning.
\newblock {\em arXiv preprint arXiv:1509.02971}.

\bibitem[Mnih et~al., 2013]{mnih2013playing}
Mnih, V., Kavukcuoglu, K., Silver, D., Graves, A., Antonoglou, I., Wierstra,
  D., and Riedmiller, M. (2013).
\newblock Playing atari with deep reinforcement learning.
\newblock {\em arXiv preprint arXiv:1312.5602}.

\bibitem[Mnih et~al., 2015]{mnih2015human}
Mnih, V., Kavukcuoglu, K., Silver, D., Rusu, A.~A., Veness, J., Bellemare,
  M.~G., Graves, A., Riedmiller, M., Fidjeland, A.~K., Ostrovski, G., et~al.
  (2015).
\newblock Human-level control through deep reinforcement learning.
\newblock {\em Nature}, 518(7540):529--533.

\bibitem[O'Donoghue et~al., 2016]{o2016combining}
O'Donoghue, B., Munos, R., Kavukcuoglu, K., and Mnih, V. (2016).
\newblock Combining policy gradient and q-learning.
\newblock {\em arXiv preprint arXiv:1611.01626}.

\bibitem[Porteus, 2002]{porteus2002foundations}
Porteus, E.~L. (2002).
\newblock {\em Foundations of stochastic inventory theory}.
\newblock Stanford University Press.

\bibitem[Scarf, 1959]{scarf1959optimality}
Scarf, H. (1959).
\newblock The optimality of (s, s) policies in the dynamic inventory problem.

\bibitem[Sutton et~al., 1998]{sutton1998introduction}
Sutton, R.~S., Barto, A.~G., et~al. (1998).
\newblock {\em Introduction to reinforcement learning}, volume 135.
\newblock MIT press Cambridge.

\bibitem[Van~Hasselt et~al., 2016]{van2016deep}
Van~Hasselt, H., Guez, A., and Silver, D. (2016).
\newblock Deep reinforcement learning with double q-learning.
\newblock In {\em Thirtieth AAAI conference on artificial intelligence}.

\end{thebibliography}

\newpage

\section*{Appendix}
\subsection*{Heuristic Ordering Policies}

Let $u_t$ is the demand of the product at period $t$. $u_t \sim \mathcal{N}(\mu_t, \sigma_t)$. $\{u_t: t \in \{0\dots N\} \}$ is a Gaussian Process determined by sets of parameters $\{\mu_t\}_{t \in \{0\dots N\}}$ and $\{\sigma_t\}_{t \in \{0\dots N\}}$.

\subsubsection*{Baseline: Ordering Mean}

Order amount is determined as follows
\begin{equation}
a_t = \hat{\mu} =  \frac{\sum_{t=0}^N \mu_t}{N} \quad \forall t
\end{equation}

\subsubsection*{Order up to $S$}
The order amount is determined by the inventory position at the end of the previous period.
\begin{equation}
a_t = (S - (I_{t-1}+a_{t-1}-u_{t-1}))^+ \quad \forall t
\end{equation}
where $S$ is determined by$
S = \Phi^{-1}(95\%)$
where $\Phi(\cdot)$ is the CDF of the Normal with mean $\hat{\mu}$ and variance $\hat{\sigma}^2$. Intuitively speaking, $S$ is the level that satisfies the demand in the period with probability 95\%. 

\end{document}